\documentclass[]{spie}  

\usepackage{booktabs}
\usepackage{amsmath,amsfonts,amssymb}
\usepackage{graphicx}
\usepackage{tikz}
\usepackage{pgfplots}
\usepackage{pgfplotstable}
\usepackage{caption}
\usepackage{subcaption}
\usepackage{floatrow}
\usepackage{multirow}

\newcommand{\xmark}{%
\tikz[scale=0.23] {
    \draw[line width=0.7,line cap=round] (0,0) to [bend left=6] (1,1);
    \draw[line width=0.7,line cap=round] (0.2,0.95) to [bend right=3] (0.8,0.05);
}}

\usepackage[colorlinks=true, allcolors=blue]{hyperref}

\title{Self-Supervised Event Representations: Towards Accurate, Real-Time  Perception on SoC FPGAs}

\author[]{Kamil Jeziorek}
\author[]{Tomasz Kryjak}
\affil[]{Embedded Vision Systems Group\\ Computer Vision Laboratory\\ Department of Automatic Control and Robotics\\ AGH University of Krakow}

\authorinfo{Further author information:\\Kamil Jeziorek: E-mail: kjeziorek@agh.edu.pl\\  Tomasz Kryjak: E-mail: tomasz.kryjak@agh.edu.pl}

\begin{document} 
\maketitle

\begin{figure}[h!]
    \centering
    \includegraphics[width=0.99\linewidth]{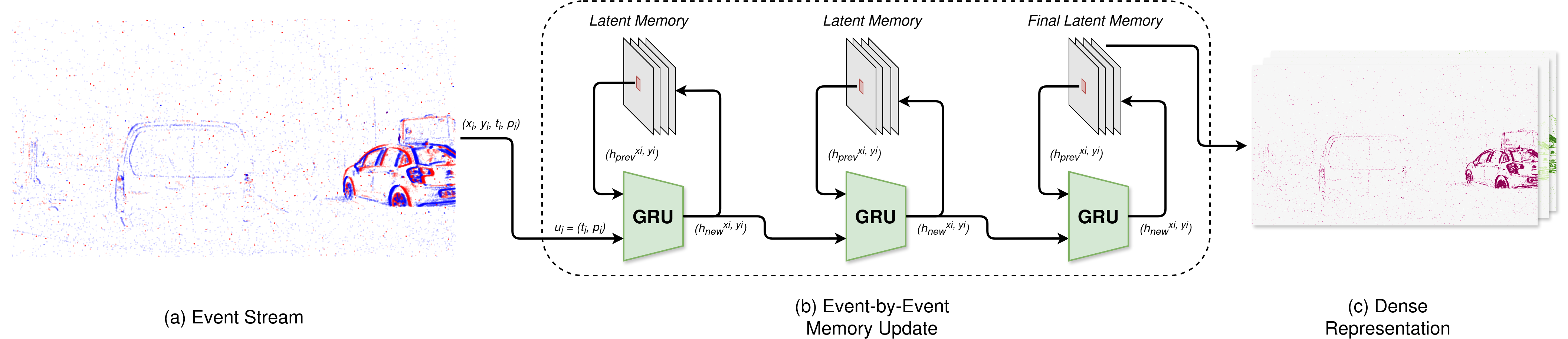}
    \caption{Overview of the proposed self‑supervised event‑representation method.  
 Sparse spatio‑temporal events (a) are processed one by one by a stack of pretrained gated‑recurrent‑unit layers (b), which update a latent‑memory tensor at the corresponding pixel $(x,y)$.  
 The latent memory has the same spatial resolution as the input and is updated continuously throughout the time window.   After all events have been processed, the final latent memory forms a dense representation (c) that preserves temporal information $t$ and polarity $p$.}
    \label{fig:teaser}
\end{figure}

\begin{abstract}

Event cameras offer significant advantages over traditional frame-based sensors. These include microsecond temporal resolution, robustness under varying lighting conditions and low power consumption. Nevertheless, the effective processing of their sparse, asynchronous event streams remains challenging. Existing approaches to this problem can be categorised into two distinct groups. The first group involves the direct processing of event data with neural models, such as Spiking Neural Networks or Graph Convolutional Neural Networks. However, this approach is often accompanied by a compromise in terms of qualitative performance. The second group involves the conversion of events into dense representations with handcrafted aggregation functions, which can boost accuracy at the cost of temporal fidelity.
This paper introduces a novel Self-Supervised Event Representation (SSER) method leveraging Gated Recurrent Unit (GRU) networks to achieve precise per-pixel encoding of event timestamps and polarities without temporal discretisation. The recurrent layers are trained in a self-supervised manner to maximise the fidelity of event-time encoding. The inference is performed with event representations generated asynchronously, thus ensuring compatibility with high-throughput sensors. The experimental validation demonstrates that SSER outperforms aggregation-based baselines, achieving improvements of 2.4\% mAP and 0.6\% on the Gen1 and 1 Mpx object detection datasets. Furthermore, the paper presents the first hardware implementation of recurrent representation for event data on a System-on-Chip FPGA, achieving sub-microsecond latency and power consumption between 1–2 W, suitable for real-time, power-efficient applications. Code is available at \href{https://github.com/vision-agh/RecRepEvent}{https://github.com/vision-agh/RecRepEvent}.

\end{abstract}


\section{INTRODUCTION}
\label{sec:intro} 

Event cameras are biologically inspired vision sensors whose operating principle differs fundamentally from that of conventional frame-based cameras \cite{Gallego_2022, 4444573, 7870263, 9063149}. In the context of an event camera, the operation of each individual pixel is characterised by independence. The detection of changes in brightness is accomplished through the implementation of a thresholding mechanism, resulting in the generation of so-called "events". Every event encodes the pixel’s spatial position, a high-precision timestamp, and the polarity of the illumination change. The asynchronous operation of pixels enables the acquisition of microsecond-level temporal resolution, while exhibiting robust performance in conditions of lighting variations with reduced power consumption of the sensor. These properties successfully overcome the key limitations of classical cameras, including low frame rates and narrow dynamic range, which can impede rapid decision-making in time-critical applications.

As the event data are sparse and asynchronous, they cannot be processed directly with methods designed for dense RGB frames. The existing literature identifies two main approaches. The first is \textbf{direct event-stream processing}, for example, using Spiking Neural Networks (SNNs) \cite{NEURIPS2021_39d4b545, 9197133, 9533514} or Graph Convolutional Neural Networks (GCNNs) \cite{Schaefer_2022_CVPR, Li_2021_ICCV, gehrig2024low}. While this enables asynchronous computation and reduced complexity, the qualitative performance often falls short of that achieved by methods based on dense representations. The second approach involves the \textbf{conversion of events into dense representations}, which are then analysed with standard Deep Convolutional Neural Networks (DCNNs) or Vision Transformers (ViTs) \cite{histograms, hats, voxel_grid, tore}. Within this paradigm, three variants prevail: (i) the aggregation of events via predefined functions (e.g. sum, maximum), (ii) the reconstruction of frames using recurrent neural networks, and (iii) the learning of end-to-end representations.

However, it should be noted that each of these solutions has certain disadvantages. Simple aggregation functions generally necessitate the quantisation of the time window, which can result in a degradation of temporal resolution or even the complete disregard of event timestamps \cite{histograms, item_ba6a35fe0743401aae4b803f132c5f05, voxel_grid}. The utilisation of U-Net–inspired architectures within recurrent networks for frame reconstruction frequently results in a substantial escalation in both computational complexity and latency \cite{Rebecq19cvpr, Scheerlinck20wacv}. Consequently, the resulting grayscale reconstructions are found to be inadequate in capturing the entirety of temporal information. Learning-based representations have been shown to adapt to event characteristics and generally yield superior results; however, they introduce additional delays, are not evaluated against the throughput of modern sensors (which are capable of hundreds of millions of events per second), and are trained jointly with the downstream model, thereby extending the overall training process \cite{Gehrig_2019_ICCV, li2022asynchronous, annamalai2022event}. The notable exception is work, where representation is selected based on the Gromov–Wasserstein Discrepancy between events and their representation \cite{Zubic_2023_ICCV}. However, this selection still depends on simple aggregation functions and inherits many of the same limitations. Despite the aforementioned points, the proposed solutions are not yet being explored with regard to hardware acceleration on SoC FPGAs or ASICs, with a view to minimising power consumption and latency.

Given the sparsity and high-throughput nature of event data, in combination with existing approaches that demonstrate gaps, the optimal representation must satisfy four criteria: (a) \textbf{event-by-event updating}, to preserve maximal temporal precision, (b) \textbf{accurate encoding of temporal information}, to leverage the fine‐grained timing information, (c) \textbf{compatibility with modern sensor throughput}, enabling processing of hundreds of millions of events per second \cite{7870263, 9063149}, (d) \textbf{hardware acceleration capability} to meet real-time requirements without excessive power consumption.

In this work, we introduce a novel method, namely \textbf{S}elf-\textbf{S}upervised \textbf{E}vent \textbf{R}epresentation (\textbf{SSER}), which is built on Gated Recurrent Unit layers \cite{gru}. In contrast to previous solutions, SSER emphasises per-pixel encoding of both event timestamps and polarities within an encoder framework, avoiding temporal discretisation. During the training phase, the encoder operates entirely in parallel, enabling the rapid adaptation of representation-generator parameters through the utilisation of the event stream alone, thus eliminating the necessity for a downstream model. During the inference process, the generation of representations can be executed in a fully asynchronous manner.  In object-detection experiments, our method outperforms aggregation-based baselines by 2.4\% mAP on the Gen1 dataset and by 0.6\% on the 1 Mpx dataset. We further demonstrate the feasibility of hardware acceleration via an SoC FPGA implementation, which achieves dynamic power consumption of 1–2 W and uses 19k–26k LUTs, 3,677–5,082 flip-flops, and 108 DSP slices per layer, depending on the configuration with 80 and 160 ns latency per event for 200 MHz and 100 MHz clock.
Our contribution can be summarised as follows:

\begin{enumerate}
    \item We introduce SSER, a novel self-supervised event-data representation using GRU networks for precise per-pixel temporal encoding.
    \item We demonstrate that SSER outperforms state-of-the-art aggregation-based methods in object detection, achieving an mAP improvement of \(0.6\%\) on the 1-Mpx dataset and \(2.4\%\) on the Gen1 dataset.
    \item We present the first SoC FPGA implementation of a recurrent event-representation generator, delivering throughput of hundreds of millions of events per second with sub-microsecond latency.
\end{enumerate}

The remainder of this paper is organised as follows. In Section \ref{sec:related}, the existing literature on event data processing is presented. The proposed method is outlined in Section \ref{sec:method}, with particular reference to its training and hardware implementation contexts. Section \ref{sec:experiments} presents a series of experiments on encoder training, detection results and utilisation of hardware implementation. The conclusion of the paper is summarised in Section \ref{sec:conclusion}.

\section{RELATED WORKS}
\label{sec:related}

In the literature, two main groups of event data processing methods can be distinguished. The first comprises methods that \textbf{operate directly on events in their original form}. Within this category, two approaches are particularly popular: Graph Convolutional Networks (GCNs)\cite{Schaefer_2022_CVPR, Li_2021_ICCV, gehrig2024low} and Spiking Neural Networks (SNNs)\cite{NEURIPS2021_39d4b545, 9197133, 9533514}. In the graph-based approach, event data are represented as graphs, where vertices correspond to individual events, and edges are established based on distance metrics. Subsequently, vertex features are updated using graph convolutions, and finally, information from the entire graph is aggregated for further analysis. In the context of Spiking Neural Networks, information processing occurs in a spike-based manner, whereby each neuron retains a membrane potential, thereby preserving the asynchronous nature of the input data. The primary benefit of both methods is the capacity for asynchronous updates, thereby significantly reducing system latency. However, despite this advantage, the performance of these methods is usually lower than that of classical vision methods, such as Convolutional Neural Networks and Transformers.

The second group of methods is centred on the \textbf{generation of dense representations} from event streams, which can subsequently be processed by ''traditional'' neural networks. Early approaches, including Event Frame \cite{item_ba6a35fe0743401aae4b803f132c5f05}, Event Count Image \cite{histograms, Zhu-RSS-18} and Surface of Active Events (SAE) \cite{6589170}, involved the representation of polarity or the timestamp of the last event, often resulting in their aggregation into separate channels. However, these methods suffer from significant loss of temporal information, particularly concerning earlier timestamps. Contemporary methods, including Exponentially Decaying Time-Surface \cite{hats, 7508476} and Time Ordered Recent Event Volume (TORE) \cite{tore}, place greater emphasis on incorporating temporal information. Nevertheless, the effectiveness of these methods is limited by manually defined aggregation functions and, in the case of the TORE method, by high computational complexity. Alternative  approaches attempt to combine several of these representation strategies \cite{9927864}, aiming to harness their individual advantages within a single framework.

Other representations, such as Voxel Grid \cite{voxel_grid} and Mixed-Density Event Stack \cite{9878430}, propose the division of the event stream into short temporal windows, the generation of local representations, and the subsequent combination of these into a larger representation that better captures the dynamics of the event stream. One recent study proposes an automatic method for selecting representation parameters based on aggregation operations (e.g. sum, variance, max, min) and data types (e.g. polarity, time, separately for positive and negative polarity) using the Gromov-Wasserstein Discrepancy measure \cite{Zubic_2023_ICCV}. Nevertheless, all these methods significantly simplify or lose precise information about event timings through aggregation within selected time sub-windows or by calculating a single summarising value for each window.

An alternative approach to generating dense representations is to utilise machine learning methods. The extant literature \cite{Gehrig_2019_ICCV, 10208507} presents examples of self-supervised learning using CNNs and Spiking-CNNs. However, in this case, the discretisation employed results in a loss of precise temporal information. Simple models based on Long Short-Term Memory (LSTM) networks, presented in \cite{annamalai2022event}, are used for noise filtering and feature extraction within short time intervals, restricting data to a two-dimensional grid only. More advanced solutions employing neural networks enable grayscale image reconstruction directly from events \cite{Rebecq19cvpr, Scheerlinck20wacv}. However, the complexity of these models limits their practical application in real-time scenarios.

The present study addresses the issue of temporal precision loss in dense event representations. The focus is on self-supervised learning of recurrent neural networks, which are optimised for hardware acceleration and high-throughput data processing.

\section{METHOD}
\label{sec:method}

The primary objective of our approach is to develop an event representation that facilitates improved preservation of temporal information. To achieve this objective, a recurrent neural network is employed, capable of asynchronously updating the feature map, thereby facilitating the creation of a dense representation in the output.

\subsection{Preliminaries}
\label{sec:prelim}

One of the unique characteristics of event cameras is their asynchronous operation. Unlike traditional cameras, which capture brightness levels across all pixels at uniform intervals, event cameras detect changes in logarithmic brightness $L$ at the individual pixel level. The generation of events in these cameras is governed by a threshold mechanism, characterised by a parameter $C$, described by the following equation:

\begin{equation}
L(x_i,y_i,t_i) - L(x_i, y_i, t_i - \Delta t_i) \geq p_i C 
\end{equation}
where $(x_i, y_i)$ represents the coordinates of a pixel, $t_i$ is the time at which the current event occurs, $t_i - \Delta t_i$ refers to the time of the last event for this pixel and $p_i \in \{-1, 1\}$ indicates the polarity of the change in brightness.
These events collectively create a spatio-temporal sequence of asynchronous data, represented as $\mathcal{E} = (e_i)_{i=0}^{N-1}$, where each event $e_i$ comprises the pixel coordinates $(x_i, y_i)$, the time of the event $t_i$, and the polarity of the event $p_i$. More details can be found in the excellent event-based vision survey \cite{Gallego_2022}.

\subsection{Recurrent Neural Networks}
\label{subsection:rnns}

Recurrent Neural Networks (RNNs) are a specific type of neural network that has been adapted for processing sequential data, such as time series. The distinguishing feature of these models is their utilisation of recurrent connections, in which the output of a sample is fed back as input to the network at the subsequent time step. The fundamental working principle of RNNs can be described as follows:

\begin{equation}
h_{t+1} = f(x_t, h_t)
\label{eq:general} 
\end{equation}
where $x_t$ represents the input vector, and $h_t$ and $h_{t+1}$ denote the hidden state vector at the current and subsequent time steps, respectively. The hidden state $h$ serves as memory, updated with each incoming input vector, enabling the capture of long-term temporal dependencies.

In the literature, three main RNN models can be distinguished. The simplest of these is the Vanilla RNN, which can be defined as follows:
\begin{equation}
h_{t+1} = \theta(W_i x_t + W_h h_t + b)
\end{equation}
where $W_i$ and $W_h$ are learned weight matrices for the input and hidden state, respectively, and the term $\theta$ is used to denote the nonlinear activation function, typically $tanh$. Due to its simplicity, this layer often suffers from the vanishing gradient problem, limiting its ability to remember distant temporal dependencies (short-term memory problem).

In order to address the issue of the vanishing gradient, the Long Short-Term Memory (LSTM)\cite{lstm} model was proposed. Each Long Short-Term Memory unit incorporates three gates and a memory cell, enabling precise control of information flow: the \emph{forget gate} ($f_t$) decides what information from the previous cell state will be discarded; the \emph{input gate} ($i_t$) determines what new information is stored in the cell; and the \emph{output gate} ($o_t$) regulates what information from the cell is passed to the hidden state. The update equations for the cell state $c_t$ and hidden state $h_t$ are as follows:
\begin{subequations}\label{eq:lstm}
\begin{align}
f_t &= \sigma(W_f x_t + U_f h_{t-1} + b_f) \\
i_t &= \sigma(W_i x_t + U_i h_{t-1} + b_i) \\
o_t &= \sigma(W_o x_t + U_o h_{t-1} + b_o) \\
\tilde{c}_t &= \tanh(W_c x_t + U_c h_{t-1} + b_c) \\
c_t &= f_t \odot c_{t-1} + i_t \odot \tilde{c}_t \\
h_t &= o_t \odot \tanh(c_t)
\end{align}
\end{subequations}
where $\sigma$ is the sigmoid function and $\odot$ denotes element-wise multiplication.

On the other hand, the \textbf{Gated Recurrent Unit} (GRU)\cite{gru} introduces simplifications compared to LSTM by utilising only two gates: the \emph{update gate} ($z_t$) decides how much of the previous hidden state is retained, and the \emph{reset gate} ($r_t$) determines how much the previous state influences the candidate hidden state calculation. The state update functions in GRU are:
\begin{subequations}\label{eq:gru}
\begin{align}
z_t &= \sigma(W_z x_t + U_z h_{t-1} + b_z) \label{eq:gru_a} \\
r_t &= \sigma(W_r x_t + U_r h_{t-1} + b_r) \label{eq:gru_b} \\
\tilde{h}_t &= \tanh\bigl(W_h x_t + r_t \odot (U_h h_{t-1} + b_h)\bigr) \\
h_t &= (1 - z_t) \odot h_{t-1} + z_t \odot \tilde{h}_t
\end{align}
\end{subequations}
The GRU merges the memory cell and hidden state, simplifying computations and maintaining the capability to model long-range temporal dependencies.

As demonstrated in Section \ref{subsection:rnns}, LSTM exhibits better performance in encoding temporal dependencies in comparison to GRU. However, it is important to note that LSTM employs two hidden states ($c_t$ and $h_t$). This characteristic is a notable disadvantage in terms of hardware implementations, necessitating an increase in data storage capacity by a factor of two. Consequently, in this study, the GRU layer was selected as the base model to achieve a balance between computational complexity and effectiveness in handling longer sequences.

Beyond these fundamental RNN models, a variety of variants have been introduced, which either simplify or extend the models \cite{8323308, zhou2016minimal, 963769}. However, the present work is specifically concerned with one variant of the GRU, namely the Minimal Gated Unit (MGU)\cite{zhou2016minimal}. In the MGU, the update gate \eqref{eq:gru_a} and reset gate \eqref{eq:gru_b}  are combined into a single forget gate ($f_t$).
\begin{subequations}\label{eq:mgu}
\begin{align}
f_t &= \sigma(W_f x_t + U_f h_{t-1} + b_f) \label{eq:mgu_a} \\
\tilde{h}_t &= \tanh\bigl(W_h x_t + f_t \odot (U_h h{t-1} + b_h)\bigr) \label{eq:mgu_b} \\
h_t &= (1 - f_t) \odot h_{t-1} + f_t \odot \tilde{h}_t \label{eq:mgu_c}
\end{align}
\end{subequations}
This variant simplifies the GRU further, facilitating simpler hardware implementations while retaining a single hidden state.

\subsection{Self-Supervised Event Representation}

The proposed method is designed to process event streams (Fig. \ref{fig:teaser}a) asynchronously. For each new event, the previous hidden state associated with the corresponding pixel is used. Let $H_0 \in \mathbb{R}^{W \times H \times C}$ denote the initialised hidden state, which simultaneously serves as the resultant representation (for last layer), where $W \times H$ specifies the sensor resolution, and $C$ the number of representation channels equal to the number of neurons in the final recurrent layer (Fig. \ref{fig:teaser}b). Upon receiving a single event $e_i = (x_i, y_i, t_i, p_i)$, the previous state value $h_{prev}^{x_i, y_i}$ is read from matrix $H_{i-1}$, and an input vector $u_i = (t_i, p_i)$ is created. The state is then updated according to the recursive function \eqref{eq:general}:
\begin{equation}
h_{new}^{x_i, y_i}= f_e(u_i, h_{prev}^{x_i, y_i}) \label{eq:general_encoder}
\end{equation}
where $f_e$ denotes the encoder update function. The new state $h_{new}^{x_i, y_i}$ is then written back into the representation matrix $H_i$ at position $(x_i, y_i)$. After processing all $n$ events within a given time window $T$, the resulting representation $H_n$ is obtained and ready for further processing by subsequent system modules (Fig. \ref{fig:teaser}c).

To train the recurrent layers, an autoencoder architecture consisting of an encoder and decoder was employed. The decoder's task is to reconstruct the original input events based on the generated representation $H_n$. For an input event sequence $\mathcal{E} = (e_0, e_1, \dots, e_n)$ and the corresponding input vectors $u_i = (t_i, p_i)$, the decoder generates event reconstructions as follows:
\begin{equation}
(d_0, d_1, \dots, d_n)= f_d(H_n, h_{prev}^{dec})
\end{equation}
where $d_i = (\hat{t}_i, \hat{p}_i)$ denotes the reconstructed timestamp and polarity values for each event, and $h^{dec}$ is the hidden state of the decoder. The training objective is to minimise the reconstruction error defined as:
\begin{equation}
\min \sum_{i = 0}^{n} |d_i - u_i| \label{eq:min_difference}
\end{equation}

\begin{figure}
    \centering
    \includegraphics[width=1\linewidth]{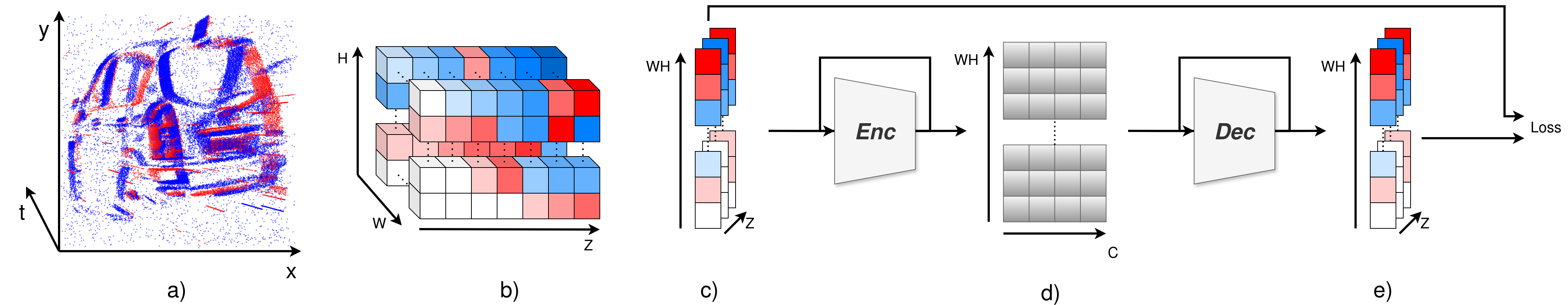}
    \caption{Overview of the training procedure for the proposed event encoder. Blue and red colours represent polarity, while shading indicates time. The spatio-temporal sequence of events (a) is first transformed into a tensor (b), where length $Z$ corresponds to the maximum number of events per pixel in the sequence. The tensor is then linearised (c), and the encoder iterates over the $Z$ timestamps, resulting in a $WH \times C$ representation (d). After this, the decoder reconstructs the events in a rolling manner based on the encoded representation (e). The loss is calculated between the input and the reconstructed events.}
    \label{fig:overview_training}
\end{figure}

In order to ensure the efficient utilisation of GPU parallelisation and to guarantee uniform input size during the training process, it is necessary to format the input data as tensors. Given an event sequence  (Fig. \ref{fig:overview_training}a) $\mathcal{E} \in \mathbb{R}^{n \times 4}$ (where $4$ represents coordinates $x, y$, time $t$, and polarity $p$), we construct a tensor $V \in \mathbb{R}^{Z \times W \times H \times 2}$, where $Z$ indicates the maximum number of events per pixel in the given time window, and $2$ corresponds to time and polarity (Fig. \ref{fig:overview_training}b). Pixels with fewer than $Z$ events are zero-padded. The tensor is then linearised into the form $V \in \mathbb{R}^{Z \times WH \times 2}$, where dimension $WH$ represents the batch size, allowing parallel processing by the encoder (Fig. \ref{fig:overview_training}c). Additionally, a mask $M \in \mathbb{R}^{Z \times WH}$ is created to indicate active (non-zero) event positions in the input tensor.

The encoder's output is a tensor $E \in \mathbb{R}^{WH \times C}$  (Fig. \ref{fig:overview_training}d), subsequently decoded iteratively. The hidden state extracted from tensor $E$ serves as the first input to the decoder's recurrent layer, with its output processed by two linear layers: the first of dimension $C \rightarrow 2$, reconstructing the event $d_i$, and the second $2 \rightarrow C$, generating input for the subsequent decoding $Z$ steps  (Fig. \ref{fig:overview_training}e).

The final reconstruction error is defined as masked mean squared error (MSE):
\begin{equation}
\text{loss} = \frac{\alpha}{ZWH}\sum_{z=0}^{Z}\sum_{w=0}^{W H}\left((V_t[z,w] - D_t[z,w])^2 M[z,w]\right) + \frac{\beta}{ZWH}\sum_{z=0}^{Z}\sum_{w=0}^{W H}\left((V_p[z,w] - D_p[z,w])^2 M[z,w]\right)
\end{equation}
where $V_t, D_t$ and $V_p, D_p$ represent input and output time and polarity, respectively. Scaling parameters $\alpha$ and $\beta$ were experimentally set to $\alpha = 1$ and $\beta = 0.1$, emphasizing accurate reconstruction of temporal information.

During the training phase, the timestamps associated with events were normalised to the range $(0, 1)$. In order to minimise the computational complexity and mitigate the influence of the "hot pixels" effect, random $64\times64$ spatial windows were selected, and the maximum sequence length $Z$ was limited to $100$. Pixels without any events were excluded from computations.

\subsection{Hardware implementation}
\label{sec:hw_impl}


The preliminary step in the hardware implementation process was model quantisation, which involved the conversion of floating-point weights and activations into low-bitwidth integer representations. Typically, two \cite{krishnamoorthi2018quantizing} approaches are applied: Post-Training Quantisation (PTQ), which involves the quantisation of a trained model based on predefined ranges of activations and weights, or Quantisation-Aware Training (QAT), which involves the simulation of quantisation during training using fake quantisation, thereby allowing the network to adjust to quantisation effects during the training phase.  In this study, we employed QAT due to its efficacy in achieving better final representation quality.

We used mixed-precision quantisation, independently adjusting precision for weights and activations, with input values (timestamp) fixed at 16 bits to ensure the precise representation of timestamps. Assuming a time window of $50$ milliseconds and microsecond-level precision, timestamp values fit within a 16-bit range ($2^{16} = 65536$).

Analysing the recurrent model equations of GRU (see Equation~\eqref{eq:gru}), five fundamental operations were identified as requiring dedicated hardware implementation: (a) \textbf{Matrix multiplication} (used in linear layers) for input vectors and previous hidden states, (b) \textbf{Element-wise vector addition}, (c) \textbf{Element-wise vector multiplication}, used in update and reset gate operations, (d) \textbf{Activation functions} $\sigma$ and $\tanh$, (e) \textbf{Subtraction} of a vector from the constant value 1.

The primary objective of the hardware implementation was to minimise latency while fully leveraging pipeline parallelism. Consequently, the modules for matrix multiplication and element-wise multiplication (based on the work \cite{wzorek2024increasing}) were fully parallelised, with parallel multipliers matching the dimension of output vectors. Linear layers of GRU defined by weight matrices $W_z$, $W_r$, $W_h$ and $U_z$, $U_r$, $U_h$ with output dimensions $d_{out}$ were combined into a single multiplier module of dimension $3 \cdot d_{out}$, whose output was subsequently divided into three parts. This structure ensures all multiplication and summation operations are completed within one clock cycle (1 CC), with an additional cycle dedicated to re-quantising the resulting vector. Element-wise multiplications were implemented in a similar manner, without the necessity of summation.

Activation functions were realised through the use of look-up tables that were precomputed with quantised activation function values, thereby ensuring minimal latency. The implementation of vector addition and subtraction was achieved through the utilisation of dedicated modules.

Each recurrent layer includes a dedicated block that stores the previous hidden states for every pixel. Its size is given by $W \times H \times d_{out} \times Precision$, where $W \times H$ represents the sensor resolution and $Precision$ denotes the number of quantisation bits. Block RAM (BRAM) modules were used for implementation.

With the proposed architecture, the processing time per event is 16 clock cycles, fully pipelined. The only architectural constraint is that two events from the same pixel must be spaced at least 16 cycles. This is due to the fact that it is not possible to process another event for the same pixel if the previous event has not been fully processed and the hidden state has not been updated. However, at a 100 MHz clock, the total latency per event is 160 ns, significantly lower than the minimal interval between consecutive events from the same pixel on typical event sensors ($\mu s$ resolution).

It is also noteworthy that the Minimal Gated Unit (see Equation~\eqref{eq:mgu}) requires the same clock cycles as a full GRU. This is due to the parallel operations of the forget and update gates in GRU, but the multipliers have a dimension of $2 \cdot d_{out}$.

\section{EXPERIMENTS}
\label{sec:experiments}

\subsection{Datasets} For our study, we used two popular automotive object detection datasets: Gen1 \cite{gen1} and 1 Mpx \cite{NEURIPS2020_c2138774}. Both consist of car-mounted recordings. 
The Gen1 dataset was captured at a resolution of \(304 \times 240\) and includes two classes -- pedestrian and car -- totalling over 255,000 object annotations and 39 hours of footage. The 1 Mpx dataset has a resolution of \(1280 \times 720\), comprises seven classes, includes 15 hours of recordings, and contains 25 million annotations.

\subsection{Encoder results}
\label{sec:encoder_results}

\begin{figure}[t]
  \centering
  \begin{subfigure}[b]{0.45\linewidth}
    \centering
    \begin{tikzpicture}
      \begin{axis}[
         tick label style={font=\scriptsize},
         height=6cm,
         label style={font=\normalsize},
         xtick=data,
         xlabel={Output size},
         ylabel={Total Loss},
         xmin=2, xmax=16,
         ymin=0, ymax=0.06,
         width=\linewidth,
         grid=major,
         cycle list name=color list
      ]
        \addplot+[mark=*, mark size=1]
          coordinates {(2,0.0398) (4,0.02966) (6,0.01782)
                       (8,0.005765) (10,0.006082) (12,0.0053276)
                       (14,0.0052402) (16,0.0044091)};
        \addplot+[mark=*, mark size=1]
          coordinates {(2,0.04556) (4,0.030934) (6,0.018827)
                       (8,0.009157) (10,0.010425) (12,0.0082284)
                       (14,0.0079174) (16,0.0065424)};
        \addplot+[mark=*, mark size=1]
          coordinates {(2, 0.05935) (4,0.03737) (6,0.03553)
                       (8,0.02771) (10,0.028534) (12,0.02697)
                       (14,0.02895) (16,0.02773)};
        \addplot+[mark=*, mark size=1, color=violet]
          coordinates {(2, 0.03403) (4,0.021758) (6,0.011504)
                       (8,0.002074) (10,0.00187) (12,0.00165)
                       (14,0.00174) (16,0.00145)};
        \legend{GRU, MGU, RNN, LSTM}
      \end{axis}
    \end{tikzpicture}
    \caption{Total loss vs.\ output size for all layers.}
    \label{fig:loss_vs_output}
  \end{subfigure}%
  \hfill
  \begin{subfigure}[b]{0.45\linewidth}
    \centering
    \begin{tikzpicture}
      \begin{axis}[
         tick label style={font=\scriptsize},
         height=6cm,
         label style={font=\normalsize},
         xtick=data,
         xlabel={Quantisation bits},
         ylabel={Total Loss},
         xmin=2, xmax=12,
         ymin=0, ymax=0.2,
         width=\linewidth,
         grid=major,
         cycle list name=color list
      ]
        \addplot+[mark=*, mark size=1]
          coordinates {(2,0.1182) (3,0.1075) (4,0.062087)
                       (5,0.02851) (6,0.016381) (7,0.0086932) (8,0.0059779) (9,0.0048934) (10,0.0056301) (11,0.0042998) (12,0.0047621)};
        \addplot+[mark=*, mark size=1]
          coordinates {(2,0.191378) (3,0.1092372) (4,0.0683028)
                       (5,0.0322229) (6,0.0181618) (7,0.0129807) (8,0.0082976) (9,0.007978) (10,0.0081449) (11,0.0081555) (12,0.0080624)};
        \legend{GRU, MGU}
      \end{axis}
    \end{tikzpicture}
    \caption{Total loss vs.\ quantisation bit for GRU and MGU}
    \label{fig:loss_vs_quantisation}
  \end{subfigure}%
  
  \caption{Ablation studies over output sizes and quantisation bit resolution for encoders.}
  \label{fig:combined}
\end{figure}
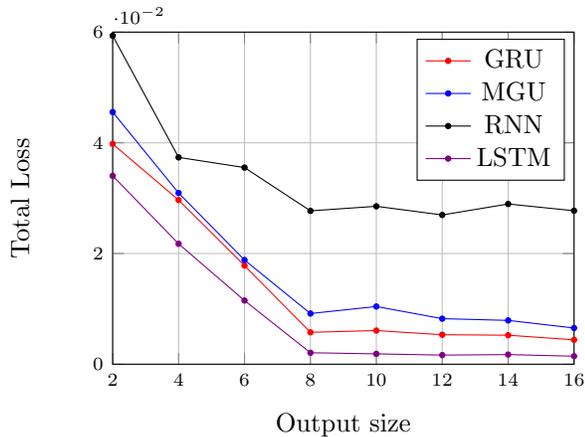
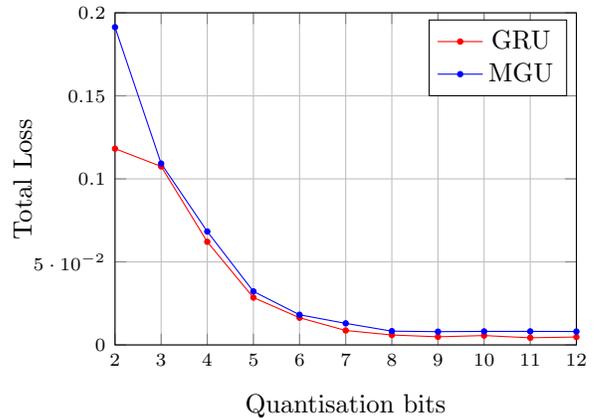

\begin{figure}[t]
    \centering
    \includegraphics[width=0.9\linewidth]{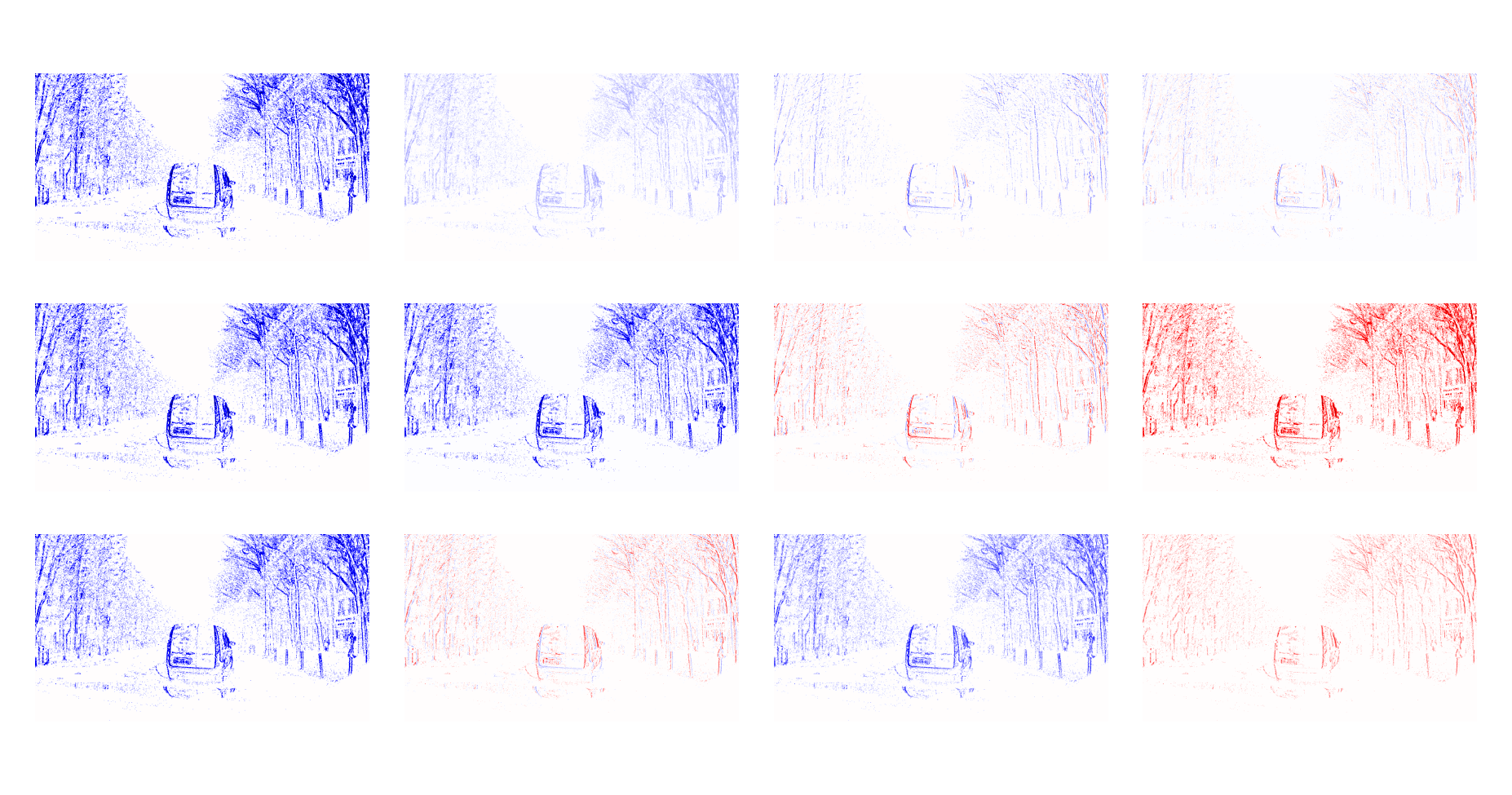}
    \caption{Example visualization of 12 GRU encoder channels on a sample from a 1-megapixel dataset. Values range from -1 to 1, with blue indicating negative values and red indicating positive values.}
    \label{fig:example_rep}
\end{figure}

In Figure \ref{fig:loss_vs_output}, we present an evaluation comparing our selected recurrent layers (GRU and MGU) with two other recurrent layers: LSTM and Vanilla RNN (hereafter referred to simply as RNN), in terms of event encoding quality. Each encoder consisted of three interconnected layers, and for each layer, the dimension of the hidden state vector was set within a range from 2 to 16. The decoder used was also a three-layer model composed entirely of GRU layers, with the number of inputs/outputs tailored to match the encoder's output.

The encoder was trained for 100 epochs using the Adam optimizer \cite{kingma2017adam}, with a learning rate of 1e-3 and weight decay of 1e-4. The training was performed on the Gen1 dataset, where each sample consisted of randomly selected events within a 200 ms temporal window, limited to 50,000 samples. The primary metric employed was the mean squared error loss between input and output events from the decoder.

According to the assumptions detailed in Section \ref{subsection:rnns}, the results indicate the effectiveness of LSTM layers in encoding event-based information, with GRU layers yielding intermediate performance between RNN and LSTM. Improvement in all model performance was observed up to approximately a hidden dimension size of 8, beyond which the decrease in loss slowed down considerably. Notably, the MGU model achieved results comparable to those obtained by the standard GRU layer (0.003 loss difference for 8 output size). This is primarily due to the moderate sequence length, making a single forget gate sufficient.

In subsequent parts of the work, the number of encoder channels was set to 12 due to comparable loss values for higher output dimensions and inspired by the number of channels used in the study presented in \cite{Zubic_2023_ICCV}.

As shown in Figure \ref{fig:loss_vs_quantisation}, we also conducted an experimental study on quantised GRU and MGU models. This experiment explored varying quantisation resolutions for the activations and weights of the recurrent layers, while keeping the input data quantised at a fixed 16-bit resolution, as described in Section \ref{sec:hw_impl}. The precision was varied between 2 and 12 bits.

The quantisation error for the GRU model using 2 bits is 0.113. Increasing the bit width further reduces this error, reaching a difference of 0.00065 at 8 bits and, for precisions above 10 bits, achieving better results than the floating-point model. For the MGU model, the discrepancies are comparable as for GRU model with a difference for 2-bit quantisation.
On this basis, an appropriate quantisation level can be selected, with a trade-off being made between accuracy and model size. The number of bits in the output representation translates directly into the memory footprint on an SoC FPGA: reducing the precision by a single bit saves $W \times H \times C$ bits, which corresponds to 24 kB for $W=H=128$ and $C=12$.

In our implementation, we applied an 8-bit quantisation to the model, as the loss curve stabilised at this value. The result of the 12-channel GRU encoder with 8-bit quantisation is shown in Figure \ref{fig:example_rep} as a visualisation of an example representation for each channel.

\subsection{Detection results}

\begin{table}[t]
    \centering
    \resizebox{0.4\textwidth}{!}{%
    \begin{tabular}{@{}lrrr@{}}
      \toprule
      Representation & mAP   & AP50  & AP75  \\
      \midrule
      2D Histogram \cite{histograms}       & 32.54 & 61.21 & 31.03 \\
      Voxel Grid \cite{voxel_grid}    & 36.63 & 65.32 & 35.10 \\
      MDES \cite{9878430}          & 35.50 & 64.21 & 33.53 \\
      TORE  \cite{tore}          & 38.68 & 67.73 & 37.06 \\
      ERGO-12 \cite{Zubic_2023_ICCV}        & 38.19 & 67.12 & 36.83 \\
      Time Surface \cite{7508476}   & 39.11 & 68.45 & 38.04 \\
      \midrule
      SSER-MGU12 (\textbf{Our})     & 39.89 & 68.91 & 38.77 \\
      SSER-GRU12 (\textbf{Our})     & \textbf{40.13} & \textbf{69.03} & \textbf{39.12} \\
      \bottomrule
    \end{tabular}
    }
    \caption{Comparison of detection performance between our Self-Supervised Event Representations vs handcrafted methods on Gen1 with ResNet-50 + YOLOX.}
    \label{tab:detection_performance}
\end{table}

For the detection task we employed the YOLOX framework\cite{ge2021yolox} (neck+head) with a ResNet‑50\cite{he2016deep} backbone whose first convolutional layer was modified to match the size of the input representation.
For the Gen1\cite{gen1} dataset, the generated representations were rescaled to a resolution of $224 \times 224$, whereas for the 1 Mpx \cite{NEURIPS2020_c2138774} dataset they were rescaled to $640 \times 640$.
The models were trained for up to 100 epochs using Stochastic Gradient Descent with Nesterov momentum \cite{pmlr-v28-sutskever13} and a cosine learning‑rate scheduler (minimum learning rate $5\times10^{-4}$, maximum learning rate $1\times10^{-2}$). The batch size was 64 for Gen1 and 32 for 1 Mpx. During training we applied random affine transformations together with random horizontal flipping. The primary evaluation metric was mean average precision (mAP).

We first analysed our approach against classical event‑camera representations (same as in work \cite{Zubic_2023_ICCV}) on the Gen1 dataset; the results are reported in Table \ref{tab:detection_performance}. Specifically, we compared against 2D Histogram \cite{histograms}, Voxel Grid \cite{voxel_grid}, Mixed‑Density Event Stack (MDES)\cite{9878430}, Time‑Ordered Recent Event Volumes (TORE)\cite{tore}, Time Surface \cite{7508476}, and Event Representation through Gromov–Wasserstein Optimization (ERGO‑12)\cite{Zubic_2023_ICCV}.

The results show that both of our methods outperform the handcrafted aggregation‑based representations in detection accuracy. The simple 2D Histogram, which ignores temporal information, performs the worst—its mAP is 7.5\% lower than that of our representations. TORE and Time Surface, which place greater emphasis on encoding event timestamps, achieve the strongest results among the competing baselines. Nevertheless, our GRU‑based representation still improves upon Time Surface by up to 1\% mAP, demonstrating both the overall superiority of our method and the beneficial effect of stronger temporal encoding on detection performance.

\begin{table}[t]
\caption{Comparison of our Self-Supervised Event Representation method on Gen1 and 1 Mpx dataset with state-of-the-art.}
\label{tab:detection_all}
\resizebox{\columnwidth}{!}{%
\begin{tabular}{@{}llcccc@{}}
\toprule
\multirow{2}{*}{Method} &
  \multirow{2}{*}{Event Repr.} &
  \multirow{2}{*}{Recurrent Representation} &
  \multirow{2}{*}{Recurrent Backbone} &
  \multicolumn{2}{c}{mAP} \\ \cmidrule(l){5-6} 
                  &                 &   &                        & Gen1 & 1 Mpx \\ \midrule
MatrixLSTM+YOLOv3 \cite{Cannici_2020_ECCV} & MatrixLSTM      & \checkmark & \multicolumn{1}{c|}{\xmark} & 31.0 & -     \\
E2VID+RetinaNet \cite{Rebecq19pami, Rebecq19cvpr}   & Reconstructions & \checkmark & \multicolumn{1}{c|}{\xmark} & 27.0 & 25.0  \\
RED \cite{NEURIPS2020_c2138774}              & Voxel Grid      & \xmark & \multicolumn{1}{c|}{\checkmark} & 40.0 & 43.0  \\
RVT-B \cite{Gehrig_2023_CVPR}            & 2D Histogram    & \xmark & \multicolumn{1}{c|}{\checkmark} & 47.2 & 47.4  \\
ASTMNet \cite{li2022asynchronous} &
  Asynchronous attention embedding &
  \xmark &
  \multicolumn{1}{c|}{\checkmark} &
  46.7 &
  48.3 \\ \midrule
Events+RetinaNet \cite{NEURIPS2020_c2138774}  & Voxel Grid      & \xmark & \multicolumn{1}{c|}{\xmark} & 34.0 & 18.0  \\
Events+SSD \cite{8594119}       & 2D Histogram    & \xmark & \multicolumn{1}{c|}{\xmark} & 30.1 & 34.0  \\
Events+RRC \cite{8575257}       & 2D Histogram    & \xmark & \multicolumn{1}{c|}{\xmark} & 30.7 & 34.3  \\
Events+YOLOv3 \cite{8793924}    & 2D Histogram    & \xmark & \multicolumn{1}{c|}{\xmark} & 31.2 & 34.6  \\ \midrule
AEGNN \cite{Schaefer_2022_CVPR}            & Graph           & \xmark & \multicolumn{1}{c|}{\xmark} & 16.3 & -     \\
EAGR \cite{gehrig2022pushing}             & Graph           & \xmark & \multicolumn{1}{c|}{\xmark} & 32.1 & -     \\
Spiking DenseNet \cite{9892618}  & Spike Train     & \xmark & \multicolumn{1}{c|}{\xmark} & 18.9 & -     \\
AsyNet \cite{messikommer2020event}           & 2D Histogram    & \xmark & \multicolumn{1}{c|}{\xmark} & 14.5 & -     \\ \midrule
\multirow{6}{*}{ResNet-50 + YOLOv6 \cite{Zubic_2023_ICCV}} &
  2D Histogram &
  \xmark &
  \multicolumn{1}{c|}{\xmark} &
  27.8 &
  23.2 \\
                  & Time Surface    & \xmark & \multicolumn{1}{c|}{\xmark} & 37.7 & 35.0  \\
                  & TORE            & \xmark & \multicolumn{1}{c|}{\xmark} & 35.9 & 34.6  \\
                  & Voxel Grid      & \xmark & \multicolumn{1}{c|}{\xmark} & 33.9 & 33.9  \\
                  & MDES            & \xmark & \multicolumn{1}{c|}{\xmark} & 34.5 & 34.2  \\
                  & EST             & \xmark & \multicolumn{1}{c|}{\xmark} & 37.0 & 34.8  \\ 
Swin-V2 + YOLOv6 \cite{Zubic_2023_ICCV}  & ERGO-12         & \xmark & \multicolumn{1}{c|}{\xmark} & 50.4 & 40.6  \\ \midrule
ResNet-50 + YOLOX (ours) &
  SSER GRU-12 &
  \checkmark &
  \multicolumn{1}{c|}{\xmark} &
  40.1 &
  35.6 \\ \bottomrule
\end{tabular}%
}
\end{table}

In Table \ref{tab:detection_all} we compare the proposed representation with previously used approaches, grouped into three categories.
The \textbf{first} category comprises models that utilise recurrent layers. Within this group, we distinguish solutions in which recurrence is limited to a single temporal window, such as MatrixLSTM and E2VID, as well as our own method. A second subgroup contains models whose backbone incorporates a recurrent component directly, thereby enabling global modelling of dependencies across successive temporal windows. This set includes RED as Convolutions + LSTM layers, RVT with Max Vision Transformers, and ASTMNet, the latter of which additionally employs an attention mechanism during generation representation.
The \textbf{second} category covers sparse methods that operate directly on events or event graphs, including AEGNN and EAGR, on spiking neural networks, or on asynchronously updated convolutional networks, termed AsyNet.
The \textbf{third} group comprises hand‑crafted representations, including Voxel Grid and 2D Histogram evaluated with RetinaNet, SSD, RRC, and YOLOv3, as well as the extensive analysis in\cite{Zubic_2023_ICCV} based on a ResNet‑50 backbone with a YOLOv6 head. As the authors did not report ERGO-12 results for ResNet-50, the values obtained with SwinTransformer-v2 are presented in that column of the table.

On the Gen1 dataset, our method surpasses MatrixLSTM by 9\% mAP and E2VID by 13\%, outperforms the best sparse method, EAGR, by 8\% mAP, and exceeds RetinaNet with a Voxel Grid by 6\% and YOLOv6 with ResNet‑50 (Time Surface) by 2.4\%. Lower scores are observed only in comparison with RVT‑B, ASTMNet, thus highlighting the benefit of recurrent backbone models when event information for static objects is absent. ERGO‑12 also outperforms our approach by more than 10\%, but this improvement derives from its Vision Transformer backbone; when the same backbone is used as in Table \ref{tab:detection_performance}, its results decrease markedly.

On the 1 Mpx dataset, the margin over E2VID widens to 10.6\% mAP; the advantage over YOLOv3+Events is 1\% and over Time Surface 0.6\%. As before, the highest results are achieved by models with a recurrent or transformer backbone (RVT‑B, ASTMNet, SwinTransformer+YOLOv6).

In summary, the proposed representation is shown to significantly outperform standard event-aggregation procedures and sparse methods. It is evident that the best performance is achieved by architectures that integrate recurrence or attention mechanisms within the backbone with a comprehensive spatio-temporal representation. This suggests that future research should focus on either adding the recurrent component deeper into the network or incorporating vision transformers.

\subsection{Hardware implementation results}

Based on earlier experiments, we implemented two recurrent‑layer models on an ZCU104 SoC FPGA: MGU and GRU, both in a 12‑channel configuration and quantised to 8‑bit precision. In each case a single layer was created together with dedicated on‑chip memory for a $128 \times 128 \times 12$ representation. Xilinx Vivado 2024.2 was utilised for the synthesis and implementation of the accelerator, from which resource utilisation and power consumption were extracted. These were then summarised in Table \ref{tab:hw_results} for two clock frequencies: 100 MHz and 200 MHz.

The representation memory occupies 48 BRAM blocks in either variant, while the GRU model requires an additional ~9k LUTs. The total utilisation of LUTs for the GRU is 22.8\%, which is 7.1\% higher than that of the MGU. In both configurations, the number of DSP blocks utilised is 108.

The static power consumption is measured at 0.6 W in both variants; however, the dynamic power consumption for GRU is approximately 30\% higher than that of the MGU.

It is crucial to emphasise that the hardware design was focused on minimising per-event latency, which was measured at 160 ns at 100 MHz and 80 ns at 200 MHz. However, it has been demonstrated that the pipeline can be relaxed from 16 to 100 clock cycles, or even further, while still achieving microsecond latency and satisfying the requirement that two events with the same pixel position are not processed concurrently. The implementation of a reduced version of the pipeline is planned for the future. This relaxation enables a substantial reduction in resource utilisation and power consumption, given that the most significant demands for LUTs and DSP blocks arise from the parallel multipliers.

\begin{table}[t]
\centering
\caption{Resource utilisation and power consumption for single recurrent layer implementation for GRU/MGU model and 100/200 MHz clock. The values presented in brackets indicate the percentage of consumption for the ZCU104 platform.}
\label{tab:hw_results}
\resizebox{\textwidth}{!}{%
\begin{tabular}{@{}llccccccc@{}}
\toprule
Clock & Model  & LUT            & FF           & BlockRAM    & DSP         & Static Power             & Dynamic Power            & Latency per event \\ \midrule
\multirow{4}{*}{100 MHz} & GRU & 26744 (11.6\%) & 5082 (1.1\%) & 0 & 108 (6.3\%) & \multirow{2}{*}{0.599 W} & \multirow{2}{*}{1.344 W} & \multirow{4}{*}{160 ns} \\
      & Memory & 25862 (11.2\%) & 0            & 48 (15.4\%) & 0           &                          &                          &                   \\ \cmidrule(lr){2-6}
      & MGU    & 19156 (8.3\%)  & 3677 (0.8\%) & 0           & 108 (6.3\%) & \multirow{2}{*}{0.597 W} & \multirow{2}{*}{1.007 W} &                   \\
      & Memory & 17115 (7.4\%)  & 0            & 48 (15.4\%) & 0           &                          &                          &                   \\ \midrule
\multirow{4}{*}{200 MHz} & GRU & 27441 (11.9\%) & 5082 (1.1\%) & 0 & 108 (6.3\%) & \multirow{2}{*}{0.607 W} & \multirow{2}{*}{2.729 W} & \multirow{4}{*}{80 ns}  \\
      & Memory & 27745 (12.0\%) & 0            & 48 (15.4\%) & 0           &                          &                          &                   \\ \cmidrule(lr){2-6}
      & MGU    & 19156 (8.3\%)  & 3677 (0.8\%) & 0           & 108 (6.3\%) & \multirow{2}{*}{0.603 W} & \multirow{2}{*}{1.986 W} &                   \\
      & Memory & 17115 (7.4\%)  & 0            & 48 (15.4\%) & 0           &                          &                          &                   \\ \bottomrule
\end{tabular}%
}
\end{table}

\section{Conclusion}
\label{sec:conclusion}

In this work, we present the first hardware implementation of a recurrent, self-supervised dense-representation generator for event data that preserves temporal information. The present study concentrated on lightweight Gated Recurrent Unit layers and a modified Minimal Gated Unit. The experimental results on the Gen1 and 1 Mpx datasets demonstrated that the proposed method exhibited higher performance in terms of detection accuracy when compared to hand-crafted event-representation techniques. The hardware prototype shows that the representation can be updated asynchronously and fully pipelined, achieving an event latency of 80–160 ns and a power envelope of 1–2 W, depending on the model and clock frequency.

Despite the fact that the proposed approach exceeds the performance of classical handcrafted representation methods, it nevertheless exhibits inferior performance in comparison to architectures that incorporate recurrent layers within a deep backbone. This underscores the pivotal function of memory mechanisms in the detection process for events, a subject which this study aims to explore in the future. In addition, the investigation of State-Space models as a potential alternative to recurrent layers is planned. Furthermore, the development of a hardware variant optimised for energy efficiency and resource utilisation is underway, with the objective of maintaining minimal per-event latency.




\acknowledgments

The work presented in this paper was supported by the National Science Centre project no. 2024/53/N/ST6/04254 entitled "F+E: Enchancing Perception through the Integration of Frame and Event Cameras" (first author) and partly supported by the programme “Excellence initiative – research university” for the AGH University of Krakow.
We also gratefully acknowledge Polish high-performance computing infrastructure PLGrid (HPC Center: ACK Cyfronet AGH) for providing computer facilities and support within computational grant no. PLG/2025/017956.
Additionally, we would like to express our gratitude to Adam Pardyl and Bartosz Zieliński for their inspiration and support during the project's initial phase.

\bibliography{report} 

\begin{thebibliography}{10}

\bibitem{Gallego_2022}
Gallego, G., Delbruck, T., Orchard, G., Bartolozzi, C., Taba, B., Censi, A.,
  Leutenegger, S., Davison, A.~J., Conradt, J., Daniilidis, K., and Scaramuzza,
  D., ``Event-based vision: A survey,'' {\em IEEE Transactions on Pattern
  Analysis and Machine Intelligence}~{\bf 44},  154–180 (Jan. 2022).

\bibitem{4444573}
Lichtsteiner, P., Posch, C., and Delbruck, T., ``A 128$\times$ 128 120 db 15
  $\mu$s latency asynchronous temporal contrast vision sensor,'' {\em IEEE
  Journal of Solid-State Circuits}~{\bf 43}(2),  566--576 (2008).

\bibitem{7870263}
Son, B., Suh, Y., Kim, S., Jung, H., Kim, J.-S., Shin, C., Park, K., Lee, K.,
  Park, J., Woo, J., Roh, Y., Lee, H., Wang, Y., Ovsiannikov, I., and Ryu, H.,
  ``4.1 a 640×480 dynamic vision sensor with a 9µm pixel and 300meps
  address-event representation,'' in [{\em 2017 IEEE International Solid-State
  Circuits Conference (ISSCC)}{\nolinebreak\hspace{0.1em}]},   66--67 (2017).

\bibitem{9063149}
Finateu, T., Niwa, A., Matolin, D., Tsuchimoto, K., Mascheroni, A., Reynaud,
  E., Mostafalu, P., Brady, F., Chotard, L., LeGoff, F., Takahashi, H.,
  Wakabayashi, H., Oike, Y., and Posch, C., ``5.10 a 1280×720 back-illuminated
  stacked temporal contrast event-based vision sensor with 4.86µm pixels,
  1.066geps readout, programmable event-rate controller and compressive
  data-formatting pipeline,'' in [{\em 2020 IEEE International Solid-State
  Circuits Conference - (ISSCC)}{\nolinebreak\hspace{0.1em}]},   112--114
  (2020).

\bibitem{NEURIPS2021_39d4b545}
Hagenaars, J., Paredes-Valles, F., and de~Croon, G., ``Self-supervised learning
  of event-based optical flow with spiking neural networks,'' in [{\em Advances
  in Neural Information Processing Systems}{\nolinebreak\hspace{0.1em}]},
  Ranzato, M., Beygelzimer, A., Dauphin, Y., Liang, P., and Vaughan, J.~W.,
  eds.,  {\bf 34},  7167--7179, Curran Associates, Inc. (2021).

\bibitem{9197133}
Gehrig, M., Shrestha, S.~B., Mouritzen, D., and Scaramuzza, D., ``Event-based
  angular velocity regression with spiking networks,'' in [{\em 2020 IEEE
  International Conference on Robotics and Automation
  (ICRA)}{\nolinebreak\hspace{0.1em}]},   4195--4202 (2020).

\bibitem{9533514}
Cordone, L., Miramond, B., and Ferrante, S., ``Learning from event cameras with
  sparse spiking convolutional neural networks,'' in [{\em 2021 International
  Joint Conference on Neural Networks (IJCNN)}{\nolinebreak\hspace{0.1em}]},
  1--8 (2021).

\bibitem{Schaefer_2022_CVPR}
Schaefer, S., Gehrig, D., and Scaramuzza, D., ``Aegnn: Asynchronous event-based
  graph neural networks,'' in [{\em Proceedings of the IEEE/CVF Conference on
  Computer Vision and Pattern Recognition (CVPR)}{\nolinebreak\hspace{0.1em}]},
    12371--12381 (June 2022).

\bibitem{Li_2021_ICCV}
Li, Y., Zhou, H., Yang, B., Zhang, Y., Cui, Z., Bao, H., and Zhang, G.,
  ``Graph-based asynchronous event processing for rapid object recognition,''
  in [{\em Proceedings of the IEEE/CVF International Conference on Computer
  Vision (ICCV)}{\nolinebreak\hspace{0.1em}]},   934--943 (October 2021).

\bibitem{gehrig2024low}
Gehrig, D. and Scaramuzza, D., ``Low-latency automotive vision with event
  cameras,'' {\em Nature}~{\bf 629}(8014),  1034--1040 (2024).

\bibitem{histograms}
Maqueda, A.~I., Loquercio, A., Gallego, G., Garcia, N., and Scaramuzza, D.,
  ``Event-based vision meets deep learning on steering prediction for
  self-driving cars,'' in [{\em 2018 IEEE/CVF Conference on Computer Vision and
  Pattern Recognition}{\nolinebreak\hspace{0.1em}]},   5419–5427, IEEE (June
  2018).

\bibitem{hats}
Sironi, A., Brambilla, M., Bourdis, N., Lagorce, X., and Benosman, R., ``Hats:
  Histograms of averaged time surfaces for robust event-based object
  classification,'' in [{\em Proceedings of the IEEE Conference on Computer
  Vision and Pattern Recognition (CVPR)}{\nolinebreak\hspace{0.1em}]},  (June
  2018).

\bibitem{voxel_grid}
Zhu, A.~Z., Yuan, L., Chaney, K., and Daniilidis, K., ``Unsupervised
  event-based learning of optical flow, depth, and egomotion,'' in [{\em
  Proceedings of the IEEE/CVF Conference on Computer Vision and Pattern
  Recognition (CVPR)}{\nolinebreak\hspace{0.1em}]},  (June 2019).

\bibitem{tore}
Baldwin, R.~W., Liu, R., Almatrafi, M., Asari, V., and Hirakawa, K.,
  ``Time-ordered recent event (tore) volumes for event cameras,'' {\em IEEE
  Transactions on Pattern Analysis and Machine Intelligence}~{\bf 45}(2),
  2519--2532 (2023).

\bibitem{item_ba6a35fe0743401aae4b803f132c5f05}
Rebecq, H., Horstschaefer, T., and Scaramuzza, D., ``Real-time visual-inertial
  odometry for event cameras using keyframe-based nonlinear optimization,'' in
  [{\em Proceedings of the British Machine Vision Conference
  (BMVC)}{\nolinebreak\hspace{0.1em}]},   16.1--16.12 (2017).

\bibitem{Rebecq19cvpr}
Rebecq, H., Ranftl, R., Koltun, V., and Scaramuzza, D., ``Events-to-video:
  Bringing modern computer vision to event cameras,'' {\em {IEEE} Conf. Comput.
  Vis. Pattern Recog. (CVPR)}  (2019).

\bibitem{Scheerlinck20wacv}
Scheerlinck, C., Rebecq, H., Gehrig, D., Barnes, N., Mahony, R., and
  Scaramuzza, D., ``Fast image reconstruction with an event camera,'' in [{\em
  {IEEE} Winter Conf. Appl. Comput. Vis.
  {(WACV)}}{\nolinebreak\hspace{0.1em}]},   156--163 (2020).

\bibitem{Gehrig_2019_ICCV}
Gehrig, D., Loquercio, A., Derpanis, K.~G., and Scaramuzza, D., ``End-to-end
  learning of representations for asynchronous event-based data,'' in [{\em
  Proceedings of the IEEE/CVF International Conference on Computer Vision
  (ICCV)}{\nolinebreak\hspace{0.1em}]},  (October 2019).

\bibitem{li2022asynchronous}
Li, J., Li, J., Zhu, L., Xiang, X., Huang, T., and Tian, Y., ``Asynchronous
  spatio-temporal memory network for continuous event-based object detection,''
  {\em IEEE Transactions on Image Processing}~{\bf 31},  2975--2987 (2022).

\bibitem{annamalai2022event}
Annamalai, L., Ramanathan, V., and Thakur, C.~S., ``Event-lstm: An unsupervised
  and asynchronous learning-based representation for event-based data,'' {\em
  IEEE Robotics and Automation Letters}~{\bf 7}(2),  4678--4685 (2022).

\bibitem{Zubic_2023_ICCV}
Zubi\'c, N., Gehrig, D., Gehrig, M., and Scaramuzza, D., ``From chaos comes
  order: Ordering event representations for object recognition and detection,''
  in [{\em Proceedings of the IEEE/CVF International Conference on Computer
  Vision (ICCV)}{\nolinebreak\hspace{0.1em}]},   12846--12856 (October 2023).

\bibitem{gru}
Cho, K., van Merrienboer, B., Bahdanau, D., and Bengio, Y., ``On the properties
  of neural machine translation: Encoder-decoder approaches,'' (2014).

\bibitem{Zhu-RSS-18}
Zhu, A., Yuan, L., Chaney, K., and Daniilidis, K., ``Ev-flownet:
  Self-supervised optical flow estimation for event-based cameras,'' in [{\em
  Proceedings of Robotics: Science and Systems}{\nolinebreak\hspace{0.1em}]},
  (June 2018).

\bibitem{6589170}
Benosman, R., Clercq, C., Lagorce, X., Ieng, S.-H., and Bartolozzi, C.,
  ``Event-based visual flow,'' {\em IEEE Transactions on Neural Networks and
  Learning Systems}~{\bf 25}(2),  407--417 (2014).

\bibitem{7508476}
Lagorce, X., Orchard, G., Galluppi, F., Shi, B.~E., and Benosman, R.~B.,
  ``Hots: A hierarchy of event-based time-surfaces for pattern recognition,''
  {\em IEEE Transactions on Pattern Analysis and Machine Intelligence}~{\bf
  39}(7),  1346--1359 (2017).

\bibitem{9927864}
Wzorek, P. and Kryjak, T., ``Traffic sign detection with event cameras and
  dcnn,'' in [{\em 2022 Signal Processing: Algorithms, Architectures,
  Arrangements, and Applications (SPA)}{\nolinebreak\hspace{0.1em}]},   86--91
  (2022).

\bibitem{9878430}
Nam, Y., Mostafavi, M., Yoon, K.-J., and Choi, J., ``Stereo depth from events
  cameras: Concentrate and focus on the future,'' in [{\em 2022 IEEE/CVF
  Conference on Computer Vision and Pattern Recognition
  (CVPR)}{\nolinebreak\hspace{0.1em}]},   6104--6113 (2022).

\bibitem{10208507}
Barchid, S., Mennesson, J., and Djéraba, C., ``Exploring joint embedding
  architectures and data augmentations for self-supervised representation
  learning in event-based vision,'' in [{\em 2023 IEEE/CVF Conference on
  Computer Vision and Pattern Recognition Workshops
  (CVPRW)}{\nolinebreak\hspace{0.1em}]},   3903--3912 (2023).

\bibitem{lstm}
Hochreiter, S. and Schmidhuber, J., ``Long short-term memory,'' {\em Neural
  Computation}~{\bf 9}(8),  1735--1780 (1997).

\bibitem{8323308}
Ravanelli, M., Brakel, P., Omologo, M., and Bengio, Y., ``Light gated recurrent
  units for speech recognition,'' {\em IEEE Transactions on Emerging Topics in
  Computational Intelligence}~{\bf 2}(2),  92--102 (2018).

\bibitem{zhou2016minimal}
Zhou, G.-B., Wu, J., Zhang, C.-L., and Zhou, Z.-H., ``Minimal gated unit for
  recurrent neural networks,'' {\em International Journal of Automation and
  Computing}~{\bf 13}(3),  226--234 (2016).

\bibitem{963769}
Gers, F. and Schmidhuber, E., ``Lstm recurrent networks learn simple
  context-free and context-sensitive languages,'' {\em IEEE Transactions on
  Neural Networks}~{\bf 12}(6),  1333--1340 (2001).

\bibitem{krishnamoorthi2018quantizing}
Krishnamoorthi, R., ``Quantizing deep convolutional networks for efficient
  inference: A whitepaper,'' {\em arXiv preprint arXiv:1806.08342}  (2018).

\bibitem{wzorek2024increasing}
Wzorek, P., Jeziorek, K., Kryjak, T., and Pinna, A., ``Increasing the
  scalability of graph convolution for fpga-implemented event-based vision,''
  {\em arXiv preprint arXiv:2411.04269}  (2024).

\bibitem{gen1}
de~Tournemire, P., Nitti, D., Perot, E., Migliore, D., and Sironi, A., ``A
  large scale event-based detection dataset for automotive,'' (2020).

\bibitem{NEURIPS2020_c2138774}
Perot, E., de~Tournemire, P., Nitti, D., Masci, J., and Sironi, A., ``Learning
  to detect objects with a 1 megapixel event camera,'' in [{\em Advances in
  Neural Information Processing Systems}{\nolinebreak\hspace{0.1em}]},
  Larochelle, H., Ranzato, M., Hadsell, R., Balcan, M., and Lin, H., eds.,
  {\bf 33},  16639--16652, Curran Associates, Inc. (2020).

\bibitem{kingma2017adam}
Kingma, D.~P. and Ba, J., ``Adam: A method for stochastic optimization,''
  (2017).

\bibitem{ge2021yolox}
Ge, Z., Liu, S., Wang, F., Li, Z., and Sun, J., ``Yolox: Exceeding yolo series
  in 2021,'' {\em arXiv preprint arXiv:2107.08430}  (2021).

\bibitem{he2016deep}
He, K., Zhang, X., Ren, S., and Sun, J., ``Deep residual learning for image
  recognition,'' in [{\em Proceedings of the IEEE conference on computer vision
  and pattern recognition}{\nolinebreak\hspace{0.1em}]},   770--778 (2016).

\bibitem{pmlr-v28-sutskever13}
Sutskever, I., Martens, J., Dahl, G., and Hinton, G., ``On the importance of
  initialization and momentum in deep learning,'' in [{\em Proceedings of the
  30th International Conference on Machine
  Learning}{\nolinebreak\hspace{0.1em}]},  Dasgupta, S. and McAllester, D.,
  eds., {\em Proceedings of Machine Learning Research} {\bf 28},  1139--1147,
  PMLR, Atlanta, Georgia, USA (17--19 Jun 2013).

\bibitem{Cannici_2020_ECCV}
Cannici, M., Ciccone, M., Romanoni, A., and Matteucci, M., ``A differentiable
  recurrent surface for asynchronous event-based data,'' in [{\em The European
  Conference on Computer Vision (ECCV)}{\nolinebreak\hspace{0.1em}]},  (August
  2020).

\bibitem{Rebecq19pami}
Rebecq, H., Ranftl, R., Koltun, V., and Scaramuzza, D., ``High speed and high
  dynamic range video with an event camera,'' {\em {IEEE} Trans. Pattern Anal.
  Mach. Intell. (T-PAMI)}  (2019).

\bibitem{Gehrig_2023_CVPR}
Gehrig, M. and Scaramuzza, D., ``Recurrent vision transformers for object
  detection with event cameras,'' in [{\em Proceedings of the IEEE/CVF
  Conference on Computer Vision and Pattern Recognition
  (CVPR)}{\nolinebreak\hspace{0.1em}]},  (2023).

\bibitem{8594119}
Iacono, M., Weber, S., Glover, A., and Bartolozzi, C., ``Towards event-driven
  object detection with off-the-shelf deep learning,'' in [{\em 2018 IEEE/RSJ
  International Conference on Intelligent Robots and Systems
  (IROS)}{\nolinebreak\hspace{0.1em}]},   1--9 (2018).

\bibitem{8575257}
Chen, N. F.~Y., ``Pseudo-labels for supervised learning on dynamic vision
  sensor data, applied to object detection under ego-motion,'' in [{\em 2018
  IEEE/CVF Conference on Computer Vision and Pattern Recognition Workshops
  (CVPRW)}{\nolinebreak\hspace{0.1em}]},   757--75709 (2018).

\bibitem{8793924}
Jiang, Z., Xia, P., Huang, K., Stechele, W., Chen, G., Bing, Z., and Knoll, A.,
  ``Mixed frame-/event-driven fast pedestrian detection,'' in [{\em 2019
  International Conference on Robotics and Automation
  (ICRA)}{\nolinebreak\hspace{0.1em}]},   8332--8338 (2019).

\bibitem{gehrig2022pushing}
Gehrig, D. and Scaramuzza, D., ``Pushing the limits of asynchronous graph-based
  object detection with event cameras,'' {\em arXiv preprint arXiv:2211.12324}
  (2022).

\bibitem{9892618}
Cordone, L., Miramond, B., and Thierion, P., ``Object detection with spiking
  neural networks on automotive event data,'' in [{\em 2022 International Joint
  Conference on Neural Networks (IJCNN)}{\nolinebreak\hspace{0.1em}]},   1--8
  (2022).

\bibitem{messikommer2020event}
Messikommer, N., Gehrig, D., Loquercio, A., and Scaramuzza, D., ``Event-based
  asynchronous sparse convolutional networks,'' in [{\em European Conference on
  Computer Vision}{\nolinebreak\hspace{0.1em}]},   415--431, Springer (2020).

\end{thebibliography}
\bibliographystyle{spiebib} 

\end{document}